\documentclass[10pt,twocolumn,letterpaper]{article}

\usepackage{cvpr}              % To produce the CAMERA-READY version
\usepackage{microtype}
\usepackage[accsupp]{axessibility}  % Improves PDF readability for those with disabilities.
\usepackage{subcaption}

\usepackage[table]{xcolor}
\definecolor{1st}{RGB}{102,194,164} % 1st
\definecolor{2nd}{RGB}{178,226,226} % 2nd
\definecolor{3rd}{RGB}{237,248,251} % 3rd
\definecolor{1stText}{RGB}{57,146,116} % For caption

\usepackage{cuted}

\usepackage{multirow}

\definecolor{cvprblue}{rgb}{0.21,0.49,0.74}
\usepackage[pagebackref,breaklinks,colorlinks,allcolors=cvprblue]{hyperref}

\title{NVGS: Neural Visibility for Occlusion Culling in 3D Gaussian Splatting}

\author{
Brent Zoomers$^{1}$\qquad
Florian Hahlbohm$^{2}$\qquad
Joni Vanherck$^{1}$\qquad
Lode Jorissen$^{1}$\\
Marcus Magnor$^{2,3}$\qquad
Nick Michiels$^{1}$\\
\small $^1$Digital Future Lab, Flanders Make, Hasselt University\qquad
\small $^2$Computer Graphics Lab, TU Braunschweig\qquad
\\\small $^3$Physics and Astronomy, University of New Mexico
}

\begin{document}
% title with teaser
\maketitle
\begin{strip}
\centering
\vspace{-1.2cm}
\includegraphics[width=1.0\linewidth]{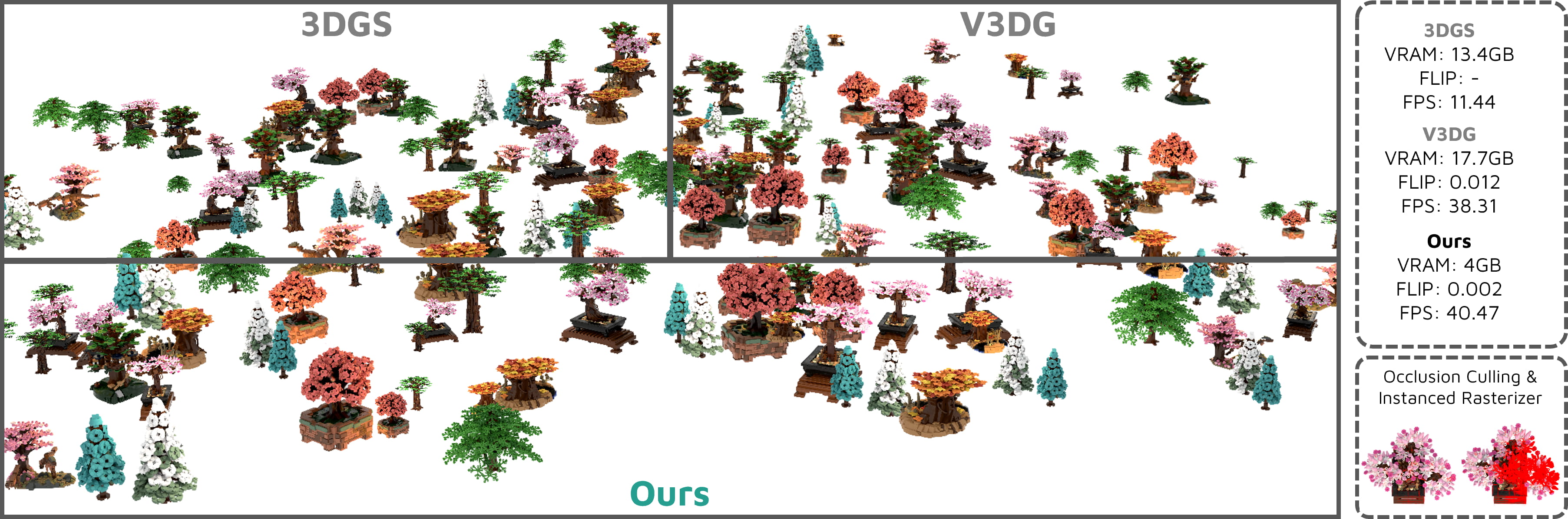}
\vspace{-0.2cm}
\captionsetup{type=figure}
\captionof{figure}{%
Shows a scene composed of multiple, separately trained 3DGS assets. \textit{Top left} shows the gsplat implementation of 3DGS~\cite{ye2025gsplat}, \textit{Top right} V3DG~\cite{Yang2025V3DG} and \textit{Bottom} Ours. Our approach uses significantly less VRAM, consistently achieves higher image quality, and increases FPS. We achieve this by combining a novel instanced rasterizer and our neural visibility MLP, which enables occlusion culling.
}\label{fig:teaser}
\end{strip}

\begin{abstract}
3D Gaussian Splatting can exploit frustum culling and level-of-detail strategies to accelerate rendering of scenes containing a large number of primitives.
However, the semi-transparent nature of Gaussians prevents the application of another highly effective technique: occlusion culling.
We address this limitation by proposing a novel method to learn the viewpoint-dependent visibility function of all Gaussians in a trained model using a small, shared MLP across instances of an asset in a scene.
By querying it for Gaussians within the viewing frustum prior to rasterization, our method can discard occluded primitives during rendering.
Leveraging Tensor Cores for efficient computation, we integrate these neural queries directly into a novel instanced software rasterizer.
Our approach outperforms the current state of the art for composed scenes in terms of VRAM usage and image quality, utilizing a combination of our instanced rasterizer and occlusion culling MLP, and exhibits complementary properties to existing LoD techniques.
The source code is available at \url{https://brent-zoomers.github.io/nvgs/}

\end{abstract}    
\section{Introduction}
\label{sec:intro}
3D Gaussian Splatting~\cite{kerbl3Dgaussians} (3DGS) has proven itself a valuable tool for 3D reconstruction in recent years, due to its ability to represent scenes with high accuracy while also facilitating both fast training and rendering.
As 3DGS transitions from research to actual applications, the need to efficiently render scenes that are growing in scale and complexity becomes increasingly important.
Previous methods~\cite{seo2024flod, Yang2025V3DG, hierarchicalgaussians24, 10993308, kulhanek2025lodge} have followed this line of thinking and introduced the traditional Level of Detail (LoD) approach to the context of 3DGS.
This enables larger scenes to be trained and rendered by utilizing hierarchical techniques to select the currently relevant parts of a scene.
Other widely used approaches in graphics, such as occlusion culling, however, are not a natural fit for 3D Gaussians due to their semi-transparent nature.
In graphics, these techniques typically rely on opaque triangles, for which visibility is straightforward to determine.
\begin{figure}[t]
\centering
\includegraphics[width=\linewidth]{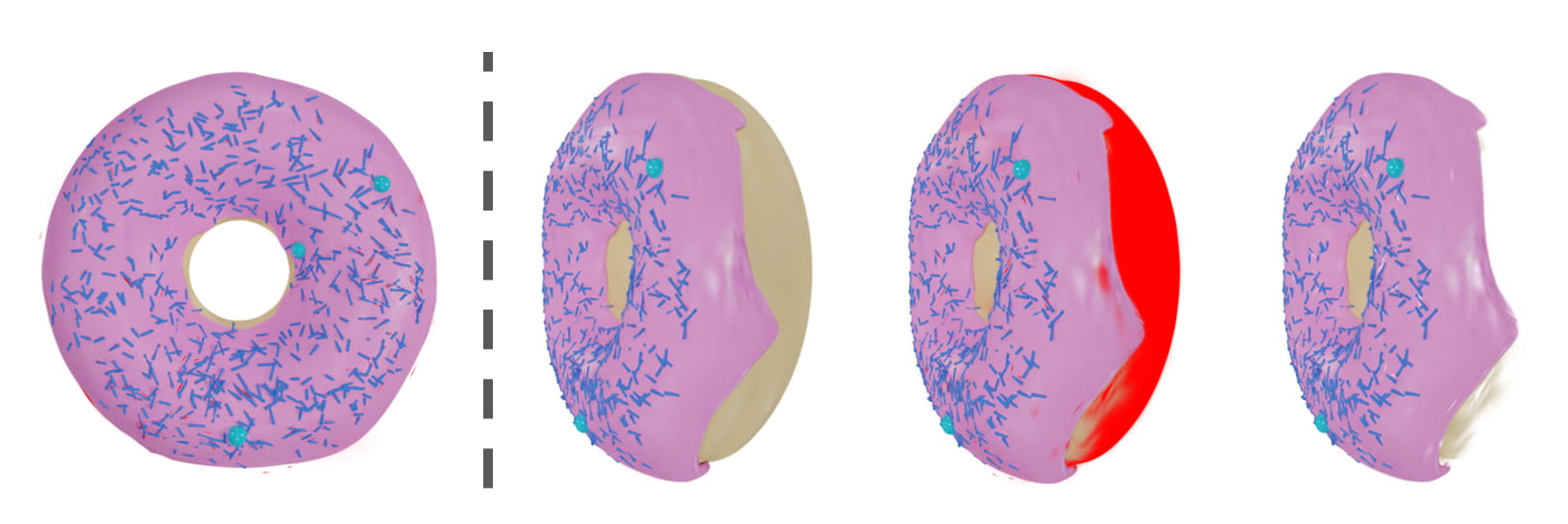}
\caption{%
\textit{Left} shows the donut asset from the front. \textit{Right} we show the donut as if rendered from the front from a different viewpoint.
We then show the Gaussians that our approach culls in red, along with what we would optimally render.}
\label{fig:bf_culling}
\end{figure}

\begin{figure}[!b]
  \centering
  \begin{subfigure}[b]{\columnwidth}
    \includegraphics[width=\linewidth]{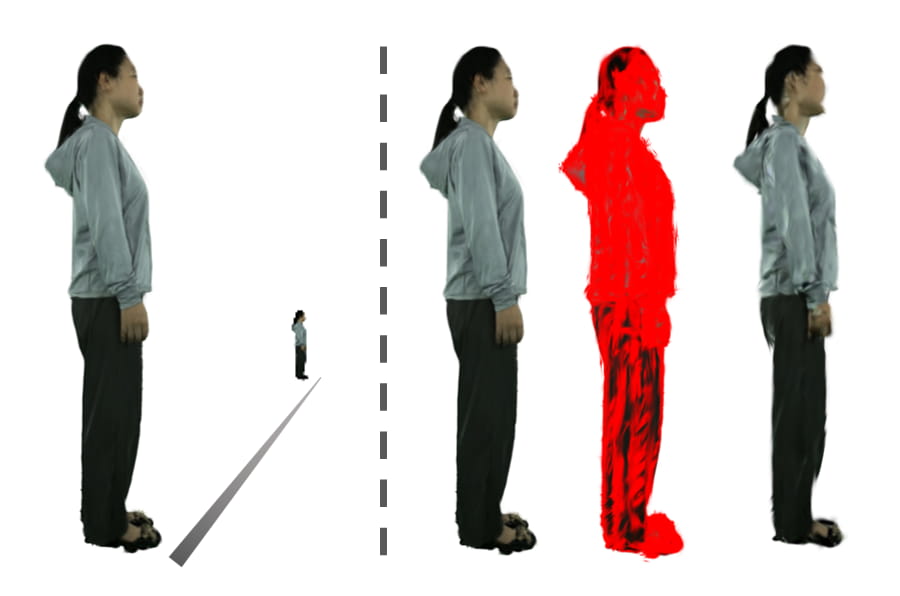}
  \end{subfigure}
  \caption{\textit{Left}: Avatar rendered at a near and far distance. \textit{Right}: We show the original render, then the Gaussians that are not used at the far distance marked in red, and on the right the Gaussians that get used at the far distance.
}
  \label{fig:lod_gt}
\end{figure}

Our results show that standard volume rendering implicitly encodes a soft form of occlusion: once transmittance saturates, subsequent splats can be discarded without any significant loss in color.
This allows us to determine Gaussians that are not visible to the current viewpoint, which, similar to backface culling for triangle meshes, are located mainly on the back side of the object, as shown in \cref{fig:bf_culling}.
Another nice property of this formulation is that it scales with distance (\cref{fig:lod_gt}).
%
% \old{As more primitives are mapped to the same set of pixels, more Gaussians can be discarded, even those lying on the front of the object.}
As more Gaussians are projected onto the same set of pixels, more will be occluded due to per-pixel transmittance saturation. This allows Gaussians on the front side to be occluded when viewed from farther away.
Capturing this information, however, requires a large amount of information to be stored.
In this work, we propose an efficient pipeline that captures and utilizes visibility information during rendering to discard Gaussians, thereby avoiding the need for expensive preprocessing in the rasterization pipeline.
We also propose a novel instanced rasterizer specifically tailored for composed scenes consisting of multiple 3DGS assets.
Gaussians are instantiated only after multiple culling phases, which significantly affects both VRAM usage and rendering speed.
Baking the visibility data into a lightweight MLP fully integrated with this rasterizer yields a highly optimized pipeline for composed scenes.
Similar to their complementary use in graphics, our occlusion culling approach shows complementary properties with LoD, opening up avenues for further optimization of large-scale scenes.
In summary, our contributions are:
\begin{itemize}
    \item We propose a novel approach to occlusion culling in 3DGS rendering that bakes the view-dependent visibility functions of all Gaussians into a small MLP. 
    \item We provide an efficient, virtualized, and instanced rendering pipeline for 3DGS capable of processing scenes over 100 million Gaussians at real-time frame rates.
    \item We demonstrate that neural queries can be integrated into the renderer to minimize excessive VRAM usage and global memory traffic, thereby improving performance.
\end{itemize}
In combination, these contributions outperform the current state of the art for composed scenes in terms of VRAM usage, image quality, and frame rate for short, medium, and, in some cases, long distances.

\section{Related Work}
\label{sec:formatting}

\subsection{3D Gaussian Splatting}
3DGS~\cite{kerbl3Dgaussians} has gained immense popularity within the community ever since its introduction.
Despite its success, 3DGS still had some limitations, some of which have been the focus of follow-up work.
To reduce the memory footprint, several approaches propose compressing the final trained scene~\cite{fan2023lightgaussian, morgenstern2024compact, navaneet2023compact3d}, using codebooks or exploiting redundancy within the Gaussian parameters.
Other works, such as StopThePop~\cite{radl2024stopthepop}, have focused on addressing popping artefacts caused by global sorting.
The densification step used in 3DGS to promote exploration inspired several follow-up works~\cite{kheradmand20243d, 10.1007/978-3-031-73036-8_20}. 
These works propose alternative strategies for splitting, cloning, or simply moving Gaussians, enabling higher-quality scene reconstructions. 
In our work, we focus on a different issue related to compression: redundancy at render time.
By baking the visibility of Gaussians into a lightweight MLP, we avoid expensive preprocessing operations on Gaussians that will not be used during rendering.

\subsection{Pruning Gaussians}
\begin{figure*}
    \centering
    \includegraphics[width=0.9\textwidth]{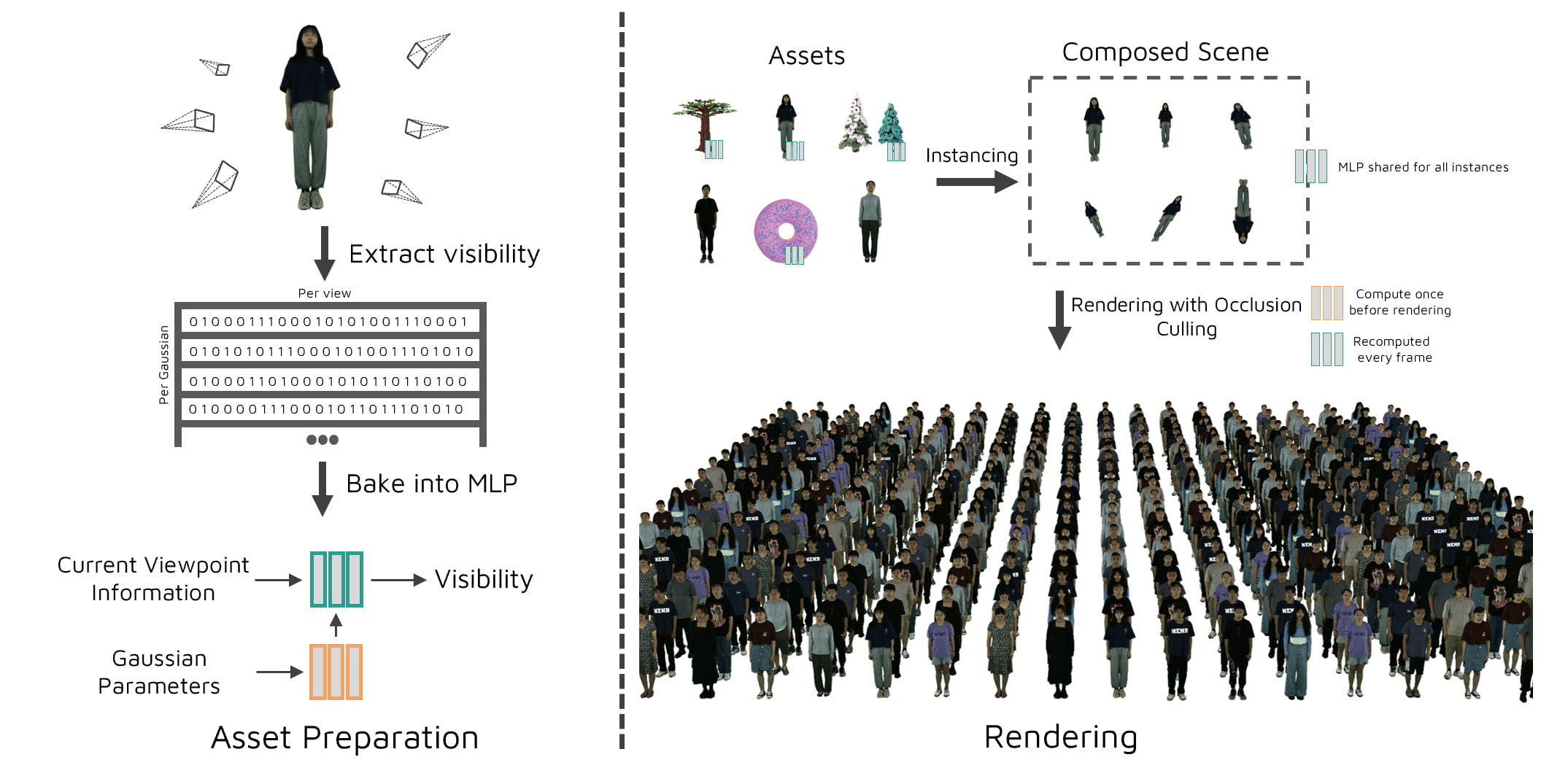}
    \caption{We extract per-Gaussian visibility by rendering to a set of training views and baking the visibility information into a lightweight MLP. Per-Gaussian parameters are encoded using a secondary MLP and precomputed once prior to rendering. At render time, our instanced rasterizer renders composed scenes containing both neural-visibility and standard 3DGS assets, leveraging neural visibility whenever available to improve efficiency by culling zero-contribution Gaussians.}%
    \label{fig:method_pipeline}
\end{figure*}
Within 3DGS, there is a high degree of redundancy, as evidenced by the extensive work on pruning and compression.
One common trend among them is their focus on global redundancy, meaning that Gaussians that do not contribute to the scene are pruned.
Zhang~\etal~\cite{zhang2024lpdgs} learns this masking by using the Gimbal-Sigmoid method to achieve an approximate binary mask, which can then be used to prune Gaussians. 
Liu~\etal~\cite{11094772} learns existence probabilities that allow gradients to flow to unused Gaussians. Gaussians with low probability of existence can be pruned both during and post-training.
Papantonakis~\etal~\cite{PapantonakisReduced3dgs} propose pruning based on spatial redundancy, where areas with overlapping, low-contributing Gaussians show potential for pruning. 
EAGLES~\cite{10.1007/978-3-031-73036-8_4} performs pruning during training, considering the contribution of Gaussians as a metric of importance.
Our method sets itself apart by not performing pruning as a global step during or after training, but rather at render time to reduce the workload per frame.
This dramatically reduces VRAM usage and increases rendering speed by avoiding expensive preprocessing of Gaussians that have no contribution to the render. 
Our approach also scales better as per-Gaussian compute increases, facilitating, \eg, more expressive shading/lighting (see \cref{sec:results}).

\subsection{Rendering of Large Scenes}
As 3DGS extends to larger, more complex scenes, the community has looked back to well-established graphics techniques for inspiration or focused on optimizing the rendering pipeline.
Both gsplat~\cite{ye2025gsplat} and Splatshop~\cite{schuetz2025splatshop} belong to the latter category, where implementation-level optimizations and approximations, \eg, radius clipping accelerate rendering of large scenes.
Level of Detail (LoD) is another viable approach, as it has historically played a crucial role in reducing the number of vertices needed to be rendered while having little or no impact on quality.
Recent work has brought LoD to the context of 3DGS, with most approaches focusing on both training and rendering of large scenes. 
H3DG~\cite{hierarchicalgaussians24} employs a divide-and-conquer strategy, where the scene is divided into distinct chunks. 
Each chunk is trained separately, after which the full scene can be rendered by efficiently consolidating the different chunks into a single hierarchy.
LODGE~\cite{kulhanek2025lodge} builds different LoD levels based on the distance of Gaussians to the camera. For far-away Gaussians, they use the 3D smoothing filter from Mip-Splatting~\cite{Yu2024MipSplatting} and prune Gaussians using their contribution as done by several other works~\cite{niemeyer2025radsplat, 10.1007/978-3-031-73036-8_4, 10943782}. Similar to other LoD approaches, they use a chunk-based rendering technique to reduce memory requirements.
Octree-GS~\cite{10993308} is based on Scaffold-GS~\cite{scaffoldgs} and uses the concept of neural anchors, which are enhanced for LoD by employing different levels of voxel grids.
The Gaussians are spawned in using the appropriate levels to achieve the desired LoD.
This, combined with a coarse-to-fine training strategy, enables the anchors to more accurately represent different levels of detail.
FLoD~\cite{seo2024flod} trains different levels of 3D Gaussian representations where each level independently learns how to represent the scene.
A level can be chosen based on hardware constraints, or multiple levels can be combined to achieve the proper LoD, with objects farther away using a level with fewer detailed Gaussians.
Recently, Liu~\etal~\cite{liu2025occlugaussian} used occlusion-aware scene partitioning for LoD by dividing a large scene into chunks.
This differs from previous works, where chunks typically follow a grid-like structure, enabling them to select Gaussians more efficiently.
Each chunk is trained individually and then consolidated again, as in the aforementioned LoD approaches.
Closest to our approach is V3DG~\cite{Yang2025V3DG}, which also focuses specifically on scenes composed of individual assets.
Their approach creates different levels of clusters based on spatial proximity.
By downsampling each cluster and optimizing it to match the original, they create a hierarchy that allows for different LoD levels.
Their approach, however, requires a large amount of VRAM and doubles storage requirements because it needs to store the different clusters.
Our approach differs from previous methods as we do not rely on LoD.
Instead, we use a method commonly used in graphics: occlusion culling.
Compared to OccluGaussian~\cite{liu2025occlugaussian}, our approach is more fine-grained, as we focus on occlusion at a Gaussian level instead of a scene level. 
The combination of our instanced rasterizer and visibility MLP significantly reduces the number of Gaussians, outperforming existing LoD approaches on VRAM usage and quality metrics.
We want to emphasize that LoD and occlusion culling are not mutually exclusive.

\section{Method}
Our approach consists of three main components.
1) Visibility extraction from a pre-trained 3D Gaussian Splatting (3DGS) asset (\cref{method:extraction}).
2) Distillation into a lightweight Multilayer Perceptron (MLP) representation, enabling efficient inference during rendering (\cref{method:baking}).
3) A novel instanced rasterizer specifically designed for 3DGS, capable of rendering scenes consisting of different assets with over 100 million Gaussians efficiently by utilizing instancing and occlusion culling (\cref{method:rendering}).
We detail each of these components in the following subsections.

\subsection{Extracting Visibility}
\label{method:extraction}
Given a trained 3DGS asset, we first extract visibility information suitable for training an MLP.
The extracted data must support unbiased training, ensuring that no periodic patterns or systematic biases are introduced.
This can result from periodicity in direction or distance sampling.
Additionally, it should remain robust to common 3DGS artifacts such as popping and aliasing, so that our approach is usable with any 3DGS-inspired method.
In most 3DGS-inspired approaches, pruning and densification are halted long before training ends.
This results in the asset containing Gaussians that are effectively useless during rendering due to their too-low opacity values.
Our first step is to remove these Gaussians from the asset to simplify training and reduce VRAM usage.
We also center the object to simplify subsequent steps and have a uniform process for all assets.
Using the means of the centered assets, we compute a per-asset minimum and maximum distance between which the camera positions will be sampled.
The minimum distance ensures that no camera is placed inside the object, while the maximum distance serves as a far bound beyond which the same quality will be displayed.
These distances are calculated based on the projected screen size of the diagonal, with the near distance set to 90\% and the far distance to 5\%.
To convert the projected screen size of an object into a distance value for the cameras, we use the following formula:
\begin{equation}
\label{eq:distance_equation}
    d = \frac{r}{\tan(\theta/2) \cdot p}
\end{equation}
where $d$ represents the distance, $r$ is half the length of the diagonal of the bounding box of the object, $p$ is the percentage of the image we want the diagonal to cover, and $\theta$ is the field of view used in the camera to create the training views.
We then uniformly sample 2000 directions using the Fibonacci sphere sampling~\cite{article_fibonacci_sphere_sampling} method.
The camera position is computed by uniformly sampling distance values between the minimum and maximum distances, then multiplying each by the direction.
To further increase robustness, specifically against projective inconsistencies, we add a random offset to the asset so that it does not appear at the center of each image.
This offset is scaled by the distance from the camera to the asset to ensure it still lies within the frustum.
Popping in 3DGS can cause Gaussians to be classified as invisible when they would be visible with a slight shift in position.
We avoid training on these artifacts by also sampling a small set of auxiliary views on the intersection of a cone rotated towards the camera and the sphere at the same distance as the camera.
Although this increases the time required for preprocessing, we demonstrate in \cref{sec:results} that it improves quality metrics.
Using this setup, a Gaussian is considered visible from a certain camera if it has a nonzero contribution to the main camera or any of the auxiliary views.
The contribution of a Gaussian G at pixel p is defined as $C_{G,p}\!=\!\alpha_p \cdot T$ where $\alpha_p$ denotes the Gaussian’s opacity at pixel $p$, and $T$ represents transmittance up to that point~\cite{kerbl3Dgaussians, 10943782,10.1007/978-3-031-73036-8_4, niemeyer2025radsplat}.

\begin{figure*}[ht]
  \centering
  \includegraphics[width=1.0\textwidth]{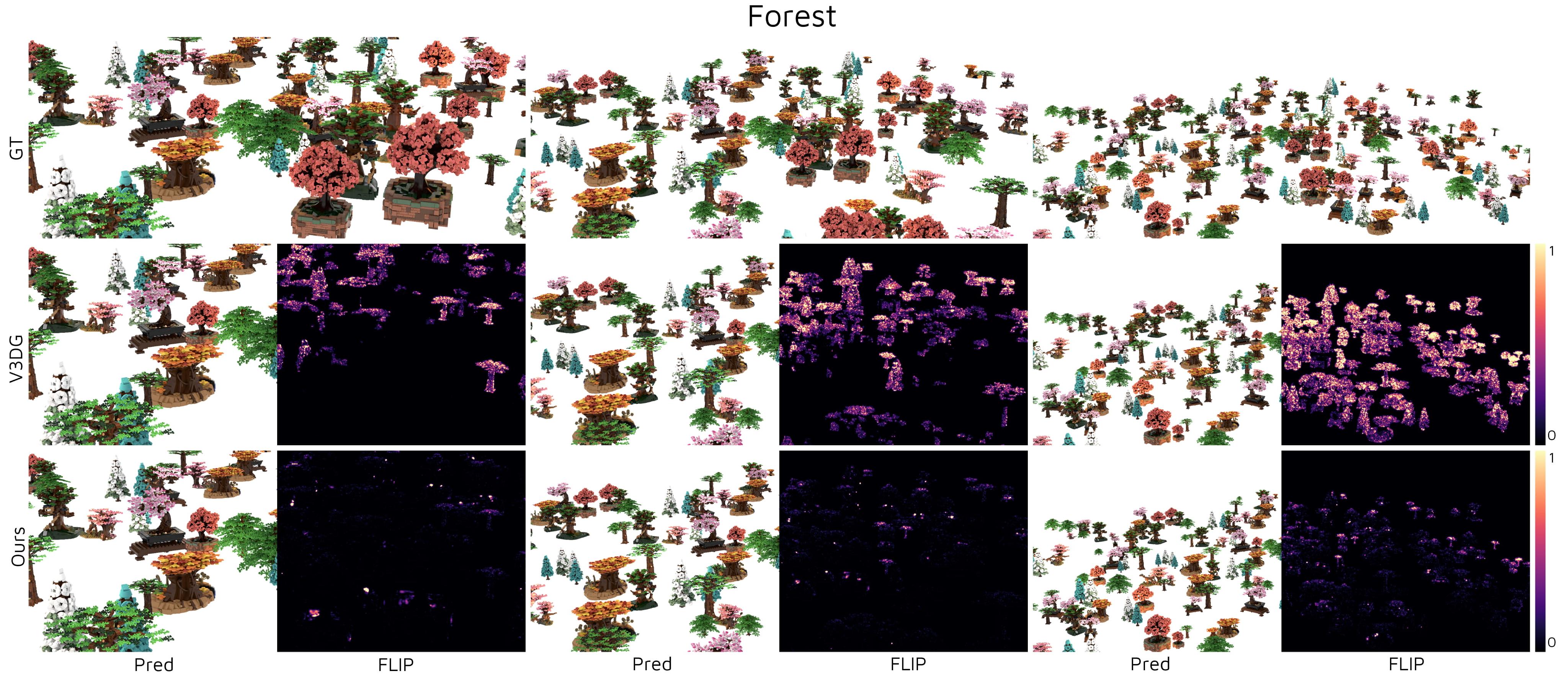}
  \caption{Shows GT render using gsplat, V3DG, and our method for close, medium, and far distances. For V3DG and ours, we show the prediction on the left and FLIP on the right. We scaled FLIP by a factor of five to facilitate better visual comparison.}
  \label{fig:visual_comparison}
\end{figure*}

\subsection{Training the Visibility MLP}
\label{method:baking}
The efficacy of our approach depends on the trade-off between the time spent querying the MLP and the time it saves during rendering.
The precomputed data in itself is too large to be used efficiently during rendering; hence, we bake it into a lightweight MLP implemented using tiny-cuda-nn (TCNN)~\cite{tiny-cuda-nn}. 
We train this MLP once per asset, after which it can be used for all instances of that asset, regardless of instance transformations or different camera properties, such as resolution or field of view (FoV).
The normalized Gaussian's mean, normalized direction, and distance to the camera, as well as the camera's forward vector and the embedded Gaussian parameters, are passed to the MLP to predict visibility.
Alongside the main visibility MLP, we train a secondary MLP to learn a six-dimensional embedding for each Gaussian based on its opacity, scaling, and rotation.
Given how frequency encoding has been used in related work to increase the expressivity of the network for similar functions~\cite{tancik2020fourfeat}, we briefly justify our decision to omit them.
While frequency encoding captures finer details, it also increases the network's input count and computation time because it uses slow trigonometric functions.
Using only five frequency bands across six inputs (mean and direction) increases the number of inputs from six to 60.
Additionally, our approach attempts to utilize Tensor Cores optimally by aligning the input dimension with a multiple of 16.
By omitting frequency encoding and using a learned feature vector for the Gaussian parameters, our approach uses exactly 16 inputs, making it optimal for the use of Tensor Cores.

\subsection{Occlusion-Aware Instanced Rendering}
\label{method:rendering}

We introduce a novel instanced rasterizer for 3DGS that enables the rendering of large composed scenes consisting of upwards of 100 million Gaussians.
Each unique asset is stored once, together with a list of instance transforms.
An instance is created only if it survives both frustum and occlusion culling, drastically reducing VRAM usage.
Before rendering, we precompute the secondary MLP once to extract the per-Gaussian feature vectors.
Per frame, we then perform frustum culling based on the Gaussian mean and decide whether to query the MLP based on the per-asset minimum distance.
The MLP input is computed per instance; hence, we must convert the inputs from global space back to the local space used during training.
For the camera's viewing direction and forward vector, we simply apply the global-to-local rotation.
The means of the Gaussians have to be translated and scaled to the local space.
For distance, we need to undo two changes.
First, we account for a potential difference in FoV between the rendering and training cameras.
Second, we undo the scaling applied to the object.
The distance in local space is then calculated as follows:
\begin{equation}
d_{t} = d_{r} \cdot \frac{f_{t}}{f_{r}} \cdot \frac{1}{s}
\end{equation} 
where the subscripts $t$ and $r$ refer to training and rendering. $d_{t}$ and $d_{r}$ represent distance, $f_{t}$ and $f_{r}$ represent the focal length, and $s$ the applied scaling factor to the asset.
Finally, we map this distance into the $[-1, 1]$ range using standard min-max normalization, based on the per-asset minimum and maximum distances.
Note how, thus far, all operations can be performed on the non-instanced Gaussians.
These inputs, together with the per-Gaussian feature vector, are passed to the MLP, which predicts per-Gaussian visibility.
A per-tile instance is created only if a Gaussian survives the previous culling steps.
From here, we use an optimized version of the original 3DGS tiled rasterization pipeline for rendering.
As shown by Lee~\etal~\cite{10763438}, the Gaussian parameters account for the largest portion of VRAM usage in 3DGS.
Hence, by avoiding the instantiation of Gaussians unless they contribute to the view, we minimize this overhead.
Furthermore, we increase rendering speed by avoiding the need to preprocess or sort discarded Gaussians.
Our approach integrates instanced rendering and occlusion culling for a highly optimized pipeline for rendering composed scenes.

\begin{figure*}[ht]
  \centering
  \includegraphics[width=0.98\textwidth]{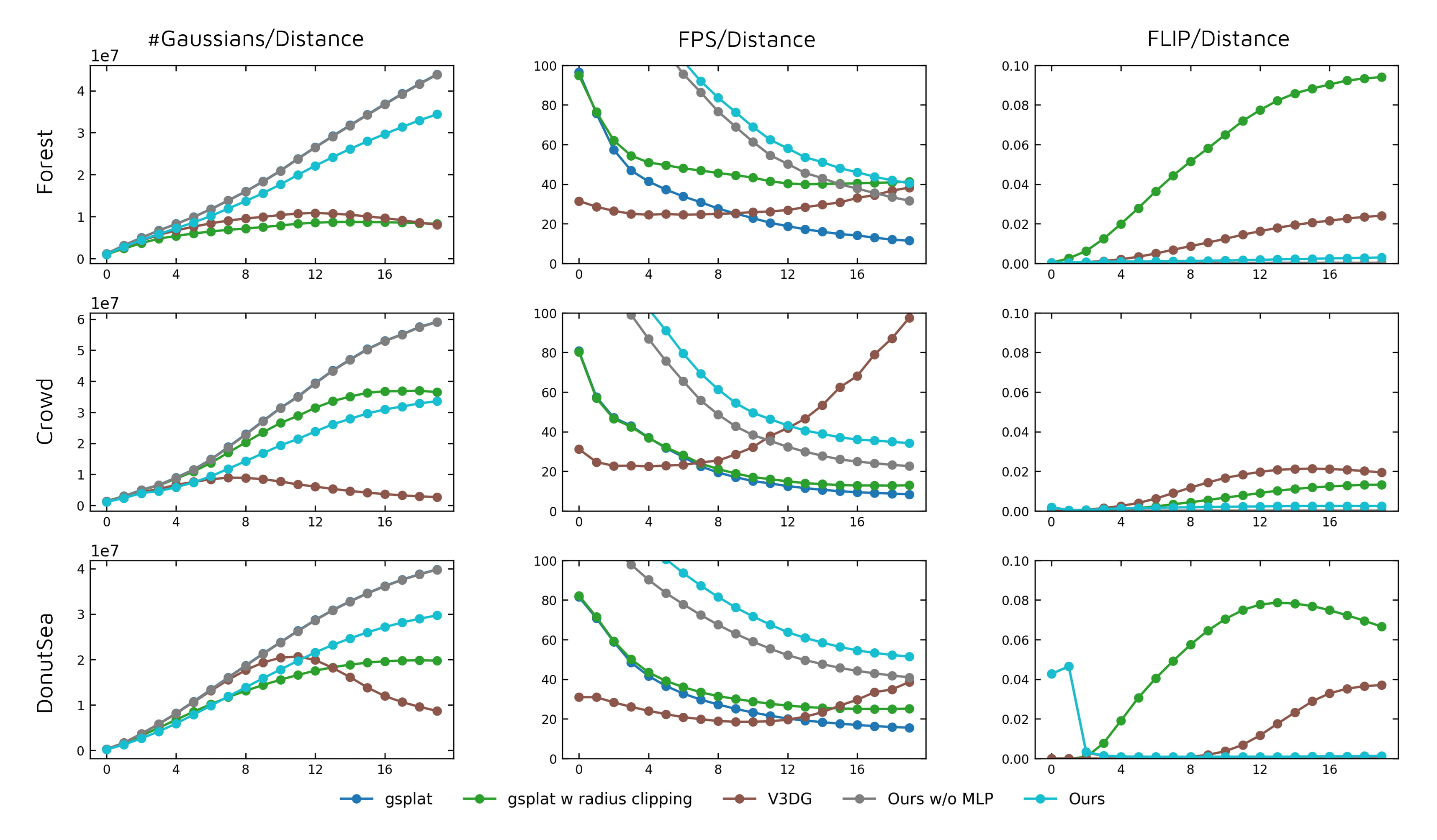}
  \caption{Shows how metrics evolve over distance for three scenes from V3DG~\cite{Yang2025V3DG}. We employ the same camera setups as in V3DG and demonstrate that, for near-to medium-range applications, our approach outperforms all other methods. For far distances, our approach remains competitive while using significantly less VRAM. We want to point out that the increase in FLIP for the \textsc{DonutSea} scene can be attributed to small numerical rounding errors when close to the near plane. This occurs for both our methods, as we do not use the gsplat rasterizer. For the number of Gaussians, using our method without MLP and gsplat yields the same result.}
  \label{fig:graph_comparison}
\end{figure*}
\section{Experiments}
\label{sec:results}
To evaluate our approach, we first compare against the state of the art for composed scenes, V3DG~\cite{Yang2025V3DG}. We also compare against gsplat and gsplat with the radius-clipping parameter enabled.
Finally, we include ablations to isolate key components of our approach.
Our novel instanced rasterizer and occlusion MLP allow our approach to compete with existing LoD methods. However, we treat LoD as complementary rather than competing. This comparison establishes the performance landscape for composed-scene pipelines and highlights the complementary aspects of both LOD and occlusion culling.
All our experiments are conducted on a single NVIDIA RTX 3090 Ti with 24 GB of VRAM. 
For comparisons with gsplat~\cite{ye2025gsplat} and V3DG~\cite{Yang2025V3DG}, we downscaled the scenes by a factor of two, ensuring that all methods could be run without exceeding VRAM limits.
We also demonstrate that our approach can run full-scale scenes without exceeding the VRAM limit.
%
%In \cref{sec:suppl} and on our project page, we provide additional videos for the full scenes and comparisons with other methods.
%
We provide additional comparisons and videos for the full scenes in the supplemental and on our project page.
\subsection{Implementation Details}
We sample 2000 views on the unit sphere using Fibonacci sampling, which yields full coverage with 5\textdegree~cones. Each point is uniformly scaled between the per-asset minimum and maximum distances (see \cref{method:extraction}).
% \old{Sampling points on the sphere to capture the contribution values is achieved using Fibonacci sphere sampling~\cite{article_fibonacci_sphere_sampling}, where each point is then scaled uniformly between the minimum and maximum distance determined for each asset, as described in \cref{method:extraction}.}
%
The random offset is then applied to the object's center point, with no offset when the object is at its minimum distance.
At other distances, the offset is scaled by linearly interpolating between 0 and the ratio of the object's minimum and maximum scales.
Both MLPs have two hidden layers with 32 neurons and ReLU activations, but differ in the number of outputs (one for the visibility MLP and six for the embedding MLP). For training, we use a customized BCE loss that weights each occurrence by its frequency in the batch, and we use the Adam optimizer~\cite{article_adam} with an initial learning rate of 2e-3 that decays to 2e-4 via a scheduler.
The scheduler applies a cosine warm-up for the first 20\% of iterations, followed by an exponential decay until completion.
Batch size is $2^{19}$, with samples drawn across views and Gaussians.
For all renderings, we match the original 3DGS blending behavior~\cite{kerbl3Dgaussians}, \ie, skip fragments with $\alpha<\frac{1}{255}$ and use 1e-4 as the transmittance threshold for early stopping.

\begin{table*}[t]
\centering
\caption{\textbf{Quantitative comparisons} on three large composed scenes ($\sim$60M Gaussians each, rendered at 1080p). We omit both baseline gsplat and ours without the MLP from image metrics, as both reproduce the ground-truth render. gsplat$^\dagger$ applies radius-based culling. Ours$^{\dagger\dagger}$ is on full-scale scenes, with Ours w/o MLP used as ground truth, as gsplat runs out of VRAM. Best results are highlighted in \textbf{\textcolor{1stText}{green}}.
}
\small
\setlength\tabcolsep{4.5pt}
\begin{tabular}{l|cccc|cccc|cccc}
\toprule
 &
\multicolumn{4}{c|}{\textbf{\textsc{Forest}}} &
\multicolumn{4}{c|}{\textbf{\textsc{Crowd}}} &
\multicolumn{4}{c}{\textbf{\textsc{DonutSea}}} \\
\textbf{Method} & PSNR$^\uparrow$ & SSIM$^\uparrow$ & FLIP$^\downarrow$ & VRAM$^\downarrow$ &
   PSNR$^\uparrow$ & SSIM$^\uparrow$ & FLIP$^\downarrow$ & VRAM$^\downarrow$ &
   PSNR$^\uparrow$ & SSIM$^\uparrow$ & FLIP$^\downarrow$ & VRAM$^\downarrow$ \\
\midrule
gsplat                  & -- & -- & -- & 13.4GB & -- & -- & -- & 16.9GB & -- & -- & -- & 11.5GB \\
gsplat$^\dagger$        & 29.5 & 0.948  & 0.056 & ~~9.7GB & 48.0  & 0.996 & 0.006 & 13.6GB & 35.6 & 0.916 &  0.053 & ~~8.3GB \\
V3DG                    & 42.3 & 0.993 & 0.012 & 17.7GB & 43.2 & 0.991 & 0.013 & 20.4GB & 58.2  & 0.983 & 0.012  & 13.9GB \\
Ours w$\slash$o MLP     & -- & -- & -- & ~~5.1GB & --  & -- & -- & ~~6.6GB & -- & --  & --  & ~~4.2GB \\
Ours                    & \cellcolor{1st}52.7 & \cellcolor{1st} 0.999 & \cellcolor{1st}0.002 & \cellcolor{1st}~~4.0GB  & \cellcolor{1st}48.6 & \cellcolor{1st}0.999 & \cellcolor{1st}0.002 & \cellcolor{1st}~~4.5GB & \cellcolor{1st}58.3 & \cellcolor{1st}0.999  & \cellcolor{1st}0.006 & \cellcolor{1st}~~3.1GB \\
\midrule
Ours$^{\dagger\dagger}$                  & 51.6 & 0.999 & 0.003 & ~~7.2GB  & 49.2 & 0.999 & 0.003 & ~~8.1GB & 64.2 & 0.999  & 0.001 & ~~6.1GB \\
\bottomrule
\end{tabular}
\label{tab:main_comparison}
\end{table*}

\subsection{Datasets and Metrics}
\label{sec:datasets_and_metrics}
Following V3DG, we evaluate the performance of our approach using the number of Gaussians, frames per second (FPS), and FLIP error~\cite{Andersson2020}.
Furthermore, we use the standard image quality metrics PSNR and SSIM~\cite{article_SSIM}, as well as VRAM usage, for comparisons.
We present results for three scenes, for which assets were sampled from three different datasets, including eight synthetic LEGO trees from the RTMV dataset~\cite{tremblay2022rtmv}, 16 real human avatars from the MVHumanNet dataset~\cite{xiong2024mvhumannet}, and a donut asset~\cite{Vagadia2021Donut3D}.
The authors of V3DG provided several scene layouts constructed from these assets, which were trained with spherical harmonics coefficients set to zero. 
For all experiments, we used the preprocessed assets, in which Gaussians with an opacity below $\frac{1}{255}$ were pruned, and the assets were re-centered.
We use the downscaled versions of these layouts to ensure they fit within VRAM for comparisons across all methods.
To create the downscaled versions of each layout, we halved the number of instances for each asset in the scene. 

\subsection{Results}
\begin{figure}[t]
\centering
\includegraphics[width=\linewidth]{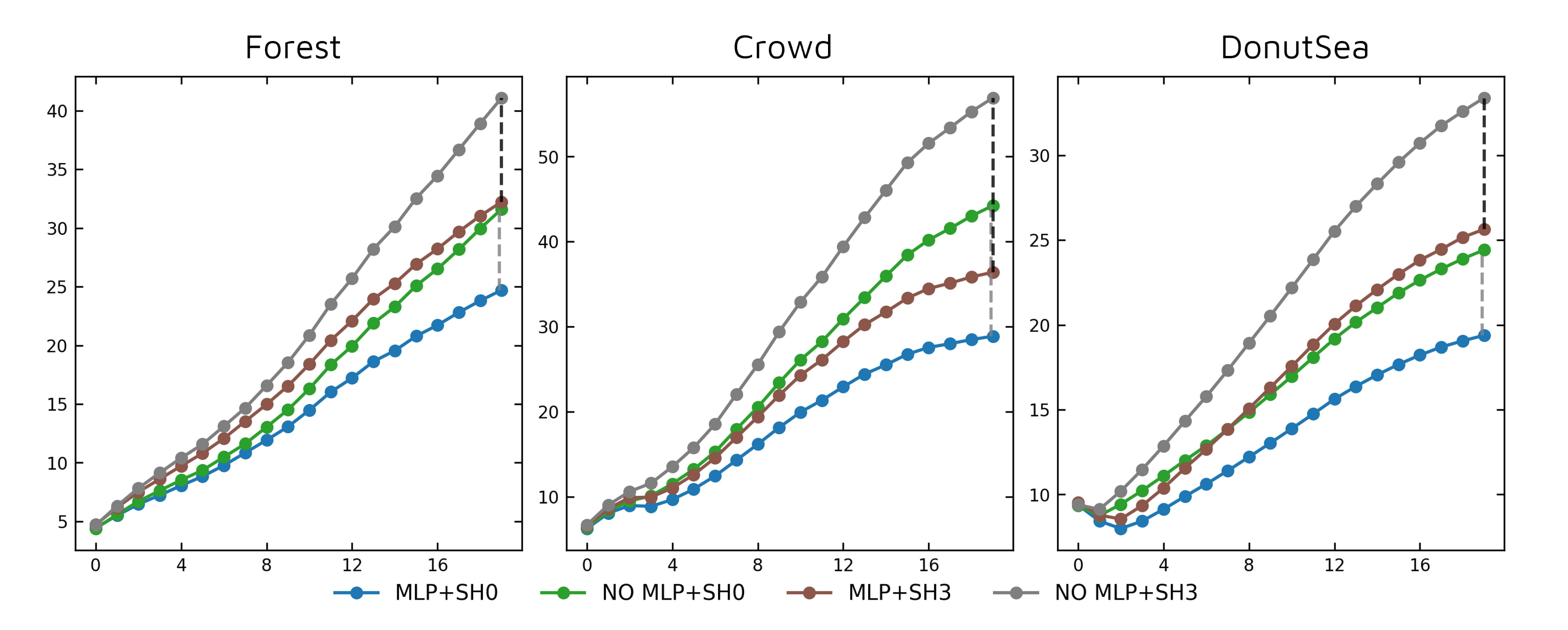}
\caption{Shows render time in milliseconds at 20 relative distances. SH represents the number of degrees used for spherical harmonics. Increasing it from 0 to 3 makes adding our MLP more beneficial, as shown by the larger vertical distance between the graphs.}
\label{fig:sh_comp}
\end{figure}

We present a qualitative visual comparison in \cref{fig:visual_comparison} comparing our approach to the reference, gsplat, and V3DG~\cite{Yang2025V3DG}.
Our method reaches near-perfect reconstruction while significantly reducing the number of Gaussians. 
% %
% The \textsc{DonutSea} scene shows almost no errors, which can be attributed to it being the smallest asset with an easy-to-learn visibility function.
% %
% The other two scenes include more Gaussians with complex visibility functions, as the leaves result in high-frequency functions, while the avatars contain the most Gaussians.
%
In \cref{fig:graph_comparison}, we show how the number of Gaussians, the FPS, and the FLIP metric change over distance. 
We want to point out that for \textsc{DonutSea} and \textsc{Forest}, the majority of Gaussians are located on the glazing of the donut ($\sim$75\%), and the crown of the trees ($\sim$66\%).
With most views taken from above, the number of Gaussians we prune using occlusion is lower than for scenes with more balanced assets, such as in \textsc{Crowd}, as demonstrated by \cref{fig:graph_comparison}.
Due to V3DG halving the number of Gaussians per level, they use fewer Gaussians, but incur the overhead of calculating which clusters to use.
For distances of low to medium range, we achieve higher rendering speeds than both V3DG and gsplat with radius clipping enabled. 
For further distances, V3DG catches up, as the difference in the number of Gaussians becomes more significant.
%
% \old{The graph clearly illustrates the complementary nature between LoD and occlusion culling as V3DG starts to increase in speed, as our approach slows down over distance.}
\Cref{fig:graph_comparison} shows the complementary nature between LoD and occlusion culling. The FPS/Distance trends show that V3DG excels at long distances, due to its drastic reduction in the number of Gaussians at those distances. Ours excels at close and medium distances by culling occluded Gaussians. Replacing the bundles needed for close distances with our MLP thus offers a simple way to integrate the two methods to exploit their strengths.
Regardless of distance, our approach outperforms gsplat with radius clipping and V3DG in terms of quality and VRAM usage.
For \textsc{DonutSea}, the initial peak in FLIP can be explained by numerical differences between our instanced rasterizer and gsplat at the near plane, as this is the only scene where the camera enters objects.
The inclusion of the MLP further increases FPS by an average of $\sim$10 over our instanced rasterizer, demonstrating the significant impact of discarding unused Gaussians.
In \cref{fig:sh_comp}, we show that the increase in rendering speed when using the MLP becomes larger as the degree of spherical harmonics used is increased. 
Besides the degree of Spherical Harmonics, other factors impact the value of our approach.
In our supplementary, we discuss several factors and how they interact with both our method and the rasterizer's properties.
As our approach works on the original asset, this allows creators to use more complex per-Gaussian calculations while still benefiting from our occlusion culling method.
\Cref{tab:build_time} shows the full training times for all datasets compared to V3DG. 
For our method, approximately 25\% of the time is spent on visibility extraction, and about 75\% on training the MLP.
The result is an MLP checkpoint that uses 18 kB, which, compared to the model's total size, occupies less than 0.1\% of the storage. 
\begin{table}[b]
\caption{\textbf{Build-time comparison} for V3DG and our method, averaged across assets. For our approach, we report combined time for visibility extraction and model training. The Trees subset contains 8 trees~\cite{tremblay2022rtmv}, MVHumanNet contains 16 humans~\cite{xiong2024mvhumannet}, and Donut contains a single object~\cite{Vagadia2021Donut3D}, with average Gaussian counts of 287965, 336554, and 44106 respectively.}
\label{tab:build_time}
\centering
\small
\setlength\tabcolsep{11.6pt}
\begin{tabular}{l|ccc}
\toprule
\textbf{Method} & Trees & MVHumanNet & Donut \\
\midrule
V3DG~\cite{Yang2025V3DG}  & 7m28s & 9m00s & \textbf{0m59s} \\
Ours  & \textbf{3m57s} & \textbf{4m11s} & 2m00s \\
\bottomrule
\end{tabular}
\end{table}
\Cref{tab:main_comparison} shows the metrics that are averaged over all distances and the peak VRAM usage. 
For base gsplat and our approach without MLP (solely the instanced rasterizer), we do not show image quality metrics as they represent the ground truth (disregarding the small numerical difference at the near plane).
The inclusion of the instanced rasterizer significantly decreases VRAM usage and increases FPS, as expected. This is a direct consequence of only instantiating Gaussians if inside the frustum. 
We use $\sim$4\texttimes\ less VRAM compared to V3DG, making our approach more scalable to larger scenes.
Using the MLP, VRAM usage is decreased further by also discarding splats inside the frustum that are occluded by others.
Besides the increase in FPS, this also reduces VRAM usage further by ~25\%.
Our approach achieves near-perfect scores across all scenes in image metrics while using significantly fewer Gaussians, demonstrating the effectiveness of our occlusion-culling MLP.

\subsection{Ablations}

\begin{figure}[t]
  \centering
\includegraphics[width=\linewidth]{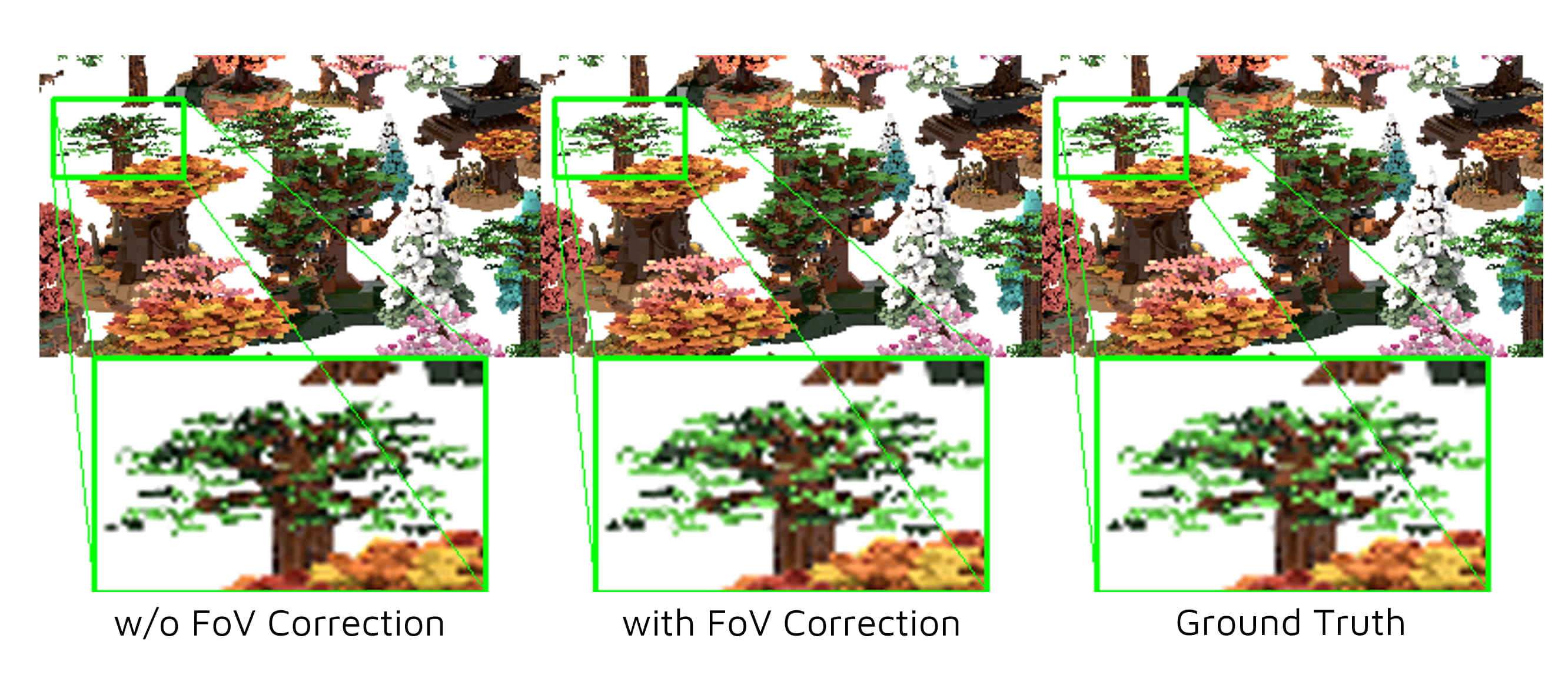}
  \caption{Shows the effect of FoV correction. Training data uses a FoV of 60\textdegree, while the rendering camera uses approximately $\sim$20\textdegree.}
  \label{fig:fov_correction}
\end{figure}
\begin{table}[b]
\caption{\textbf{Ablation study}. FPS is measured at the max distance to isolate the impact for far away viewpoints. FLIP is averaged over all distances, VRAM shows peak usage. We optionally use radius clipping to increase rendering speed, albeit at the cost of quality.}
\centering
% \small
\setlength\tabcolsep{7pt}
\begin{tabular}{l|ccc}
\toprule
\textbf{Method} & FPS$^\uparrow$ &  FLIP$^\downarrow$ & VRAM$^\downarrow$ \\
\midrule
LongLat sampling                        & 61.04   & 0.0170    & 2.64GB \\
Fibonacci Sampling                      & 61.61   & 0.0168    & 2.63GB \\
+ Random Offset                         & 59.75   & 0.0142    & 2.72GB \\
+ Auxiliary Views                       & 53.23   & 0.0113    & 2.98GB  \\
Full (+ FoV Correction)          & 42.02   & 0.0031    & 3.88GB \\
+ Radius Clipping                          & 50.59   & 0.0067    & 3.79GB \\
\bottomrule
\end{tabular}
\label{tab:ablation}
\end{table}

In \cref{tab:ablation}, we show the impact of different additions to our approach.
A first small improvement results from replacing Longitude-Latitude sampling with Fibonacci sphere sampling.
Including the random offset and adding auxiliary views increases VRAM usage and decreases FPS.
Both additions, however, serve as robustness measures against small artifacts, leading to more Gaussians being correctly predicted as visible, as reflected in the FLIP metric.
In the case of auxiliary views, it helps against training on popping artifacts, while the random offset helps with projection artifacts.
We argue that these overpredictions are preferred over worse visual results, even if they reduce FPS and increase VRAM usage.
Applying the FoV correction yields the largest change across all metrics.
Given that training was done using a camera with a FoV of 60\textdegree, and the layouts in V3DG all use lower FoV, the perceived distance from the camera to the Gaussian was consistently under-estimated.
This leads the model to predict visibility as if the object were viewed from farther away.
Lastly, we optionally apply radius clipping to accelerate the rendering of distant distances.
In gsplat, radius clipping works by using the largest radius of the axis-aligned bounding box to decide whether a Gaussian should be pruned.
We instead use the determinant of the 2D covariance matrix, which corresponds to the 2D area at $1\sigma$.
Including radius clipping increases the FLIP error, but allows for faster rendering at far distances.

\subsection{Limitations and Future Work}
While our method significantly improves VRAM efficiency for rendering 3DGS scenes, it is not without its limitations.
The performance of the MLP depends on the number of Gaussians and the complexity of their visibility functions, which may require additional overhead to ensure robustness for complex objects.
We have demonstrated the efficacy of our method on individual assets, but the same properties hold true for full scenes.
Using different sampling or scene-division strategies offers interesting avenues for further research.
% We have demonstrated the efficacy of our visibility MLP on individual assets, but it can be extended to full scenes using camera-sampling strategies that account for scene geometry.
A combination of occlusion culling and LoD could broaden the range of renderable scenes, as it does for mesh-based scenes.
Lastly, engineering optimizations, \eg, CUDA Streams, may further optimize the efficacy of our approach.
We leave these points as future work.

\section{Conclusion}
\label{sec:conclusion}
In this work, we presented a novel object-level occlusion culling approach for 3DGS.
By encoding visibility in a small MLP, we enable efficient rendering by discarding Gaussians, avoiding the need for expensive preprocessing for a significant number of Gaussians, which directly addresses a key bottleneck in prior approaches.
We also introduced a novel instanced 3DGS rasterizer that integrates MLP evaluation into the rendering pipeline, enhancing performance while reducing VRAM requirements.
Combined, these contributions enable the efficient, real-time rendering of composed scenes with up to 100 million Gaussians.
By introducing a pipeline specifically designed for composed scenes, our work lowers the barrier to rendering large-scale composed scenes for, \eg, gaming or movie industry applications.

\section*{Acknowledgements}
This research was partly funded by the FWO fellowship grants (1SHDZ24N, 1S80926N), the Special Research Fund (BOF) of Hasselt University (R-14360, R-14436), the NORM.AI SBO
project, the DFG project ``Real-Action VR'' (ID 523421583), and the Flanders Make SBO project XRTWin (R-12528). This work was made possible with the support of MAXVR-INFRA, a scalable and flexible infrastructure that facilitates the transition to digital-physical work environments.

% \clearpage
% References
% \clearpage

{
    \small
    \bibliographystyle{ieeenat_fullname}
    \bibliography{main}
}

\appendix
\maketitlesupplementary
\section{Supplementary}
\label{sec:suppl}

This supplementary document provides additional results and evaluations for all layout files, including renders and FLIP~\cite{Andersson2020} comparisons for each layout.
This includes comparisons for the full layouts from V3DG~\cite{Yang2025V3DG}, where we use our instanced rasterizer without the MLP for reference renderings.
Results for individual assets are presented, along with visual comparisons of the ground-truth contributions and our predicted values.
We also encourage the reader to view the supplementary videos, as the results can be more clearly inspected when the camera is in motion.

\subsection{Per-asset Results}
In \cref{tab:gaussian_stats} we show per-asset results.
These metrics were measured for a circular camera trajectory around each asset.
For the assets in the \textsc{DonutSea} and \textsc{Crowd} layouts, our MLP culls a significant number of Gaussians, ranging from $\sim$29\% to $\sim$40\%.
For the trees from the \textsc{Forest} layout, the number of Gaussians that can be pruned depends strongly on the crown structure and trunk thickness.
In the cases of oak and dinosaur, we show in the video that these exact factors complicate the accurate learning of the visibility function.
On average, our approach increases the percentage of useful Gaussians passed to the rasterizer while significantly decreasing the total number of Gaussians processed.
Notably, these Gaussians can be discarded without loss of quality, as they do not contribute to the ground truth renders.

\begin{figure}[b]
  \centering
  % --- First Subfigure (GT) --
  \begin{subfigure}[b]{0.49\columnwidth}
    \centering
    \includegraphics[width=0.6\linewidth]{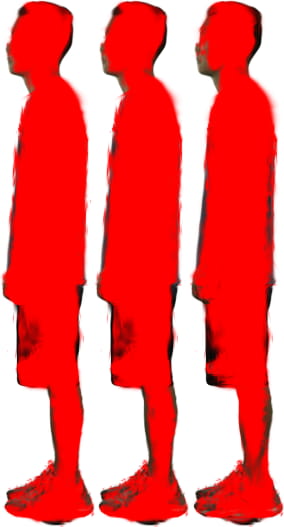}
    \caption{100843}
  \end{subfigure}
    \begin{subfigure}[b]{0.49\columnwidth}
    \centering
    \includegraphics[width=0.75\linewidth]{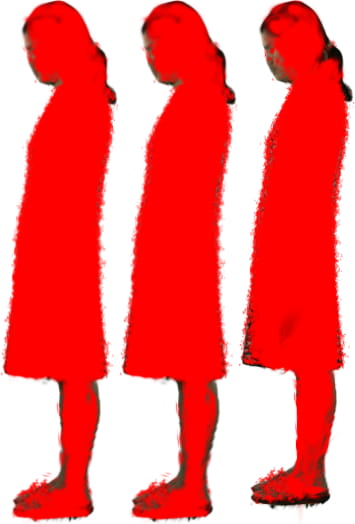}
    \caption{100845}
  \end{subfigure}
  \caption{From left to right, we show the ground truth zero-contribution Gaussians, ground truth zero-contribution using 6 auxiliary views, and what the MLP predicts. The two chosen assets are the best- and worst-performing assets from the MVHumanNet~\cite{xiong2024mvhumannet} dataset in terms of Gaussian pruning percentage.}
  \label{fig:gtpred0}
\end{figure}

\begin{figure}[b]
  \centering
  % --- First Subfigure (GT) --
  \begin{subfigure}[b]{1.0\columnwidth}
    \centering
    \includegraphics[width=\linewidth]{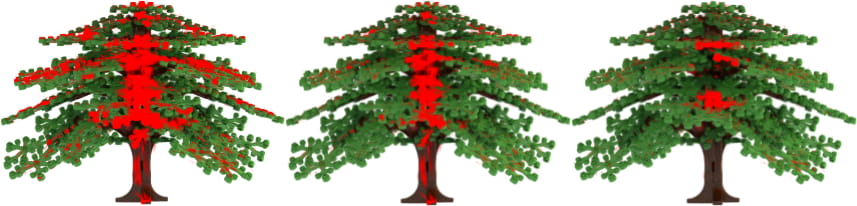}
    \caption{oak}
  \end{subfigure}
    \begin{subfigure}[b]{1.0\columnwidth}
    \centering
    \includegraphics[width=\linewidth]{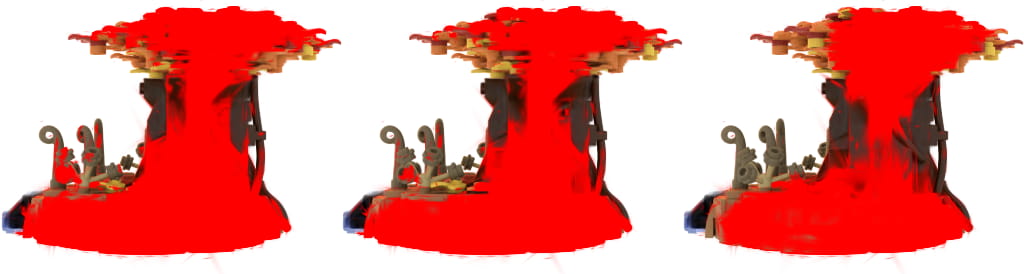}
    \caption{fall}
  \end{subfigure}
  \caption{From left to right, we show the ground truth zero-contribution Gaussians, ground truth zero-contribution using 6 auxiliary views, and what the MLP predicts. The two chosen assets are the best- and worst-performing assets from the RTMV~\cite{tremblay2022rtmv} trees subset in terms of Gaussian pruning percentage.}
  \label{fig:gtpred1}
\end{figure}

\begin{figure}[t]
  \centering
  % --- First Subfigure (GT) --
  \begin{subfigure}[b]{1.0\columnwidth}
    \centering
    \includegraphics[width=\linewidth]{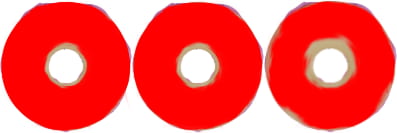}
    \caption{donut}
  \end{subfigure}

  \caption{From left to right, we show the ground truth zero-contribution Gaussians, ground truth zero-contribution using 6 auxiliary views, and what the MLP predicts. We show this for the donut asset~\cite{Vagadia2021Donut3D}.}
  \label{fig:gtpred2}
\end{figure}

In \cref{fig:gtpred0,fig:gtpred1,fig:gtpred2}, we present side-by-side images of a selection of assets, comparing the ground truth selection with the occlusions our MLP predicts.
From left to right, we show the ground truth in the single view, the ground truth we actually learn using a small set of auxiliary views to counter training on popping artifacts, and what our MLP predicts as occluded.
In all views, (predicted) non-contributing splats are rendered in red, as if the camera were on the opposite side of the asset.
For the avatars and trees, we display the asset with the best and worst performance in terms of the percentage of Gaussians pruned.
For the trees, notice that the two that perform worse have many small gaps in their crowns, resulting in a higher frequency of visibility changes compared to the avatars or donut.
Including frequency encodings and avoiding auxiliary views both increase the number of Gaussians we predict as occluded, at the cost of rendering speed and robustness against popping artifacts due to harder-to-learn visibility functions.
Depending on the application, the selection of process steps can be fine-tuned; however, our proposed method, with all steps enabled, achieves consistent results across assets and datasets.

\subsection{Large-scale Scenes}
\begin{figure*}[t]
  \centering
  \begin{subfigure}[b]{1.0\textwidth}
    \centering
    \caption*{\textsc{DonutSea}}
    \includegraphics[width=1.0\textwidth]{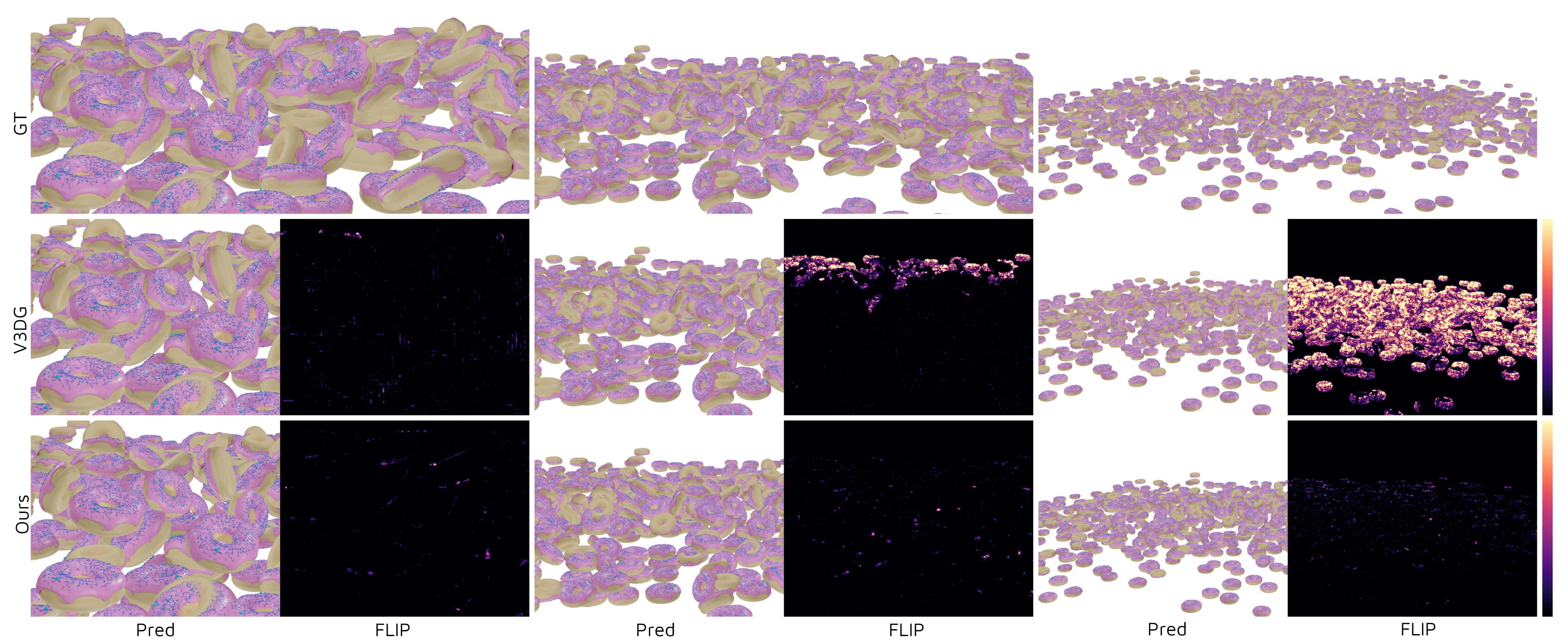}
  \end{subfigure}
  \begin{subfigure}[b]{1.0\textwidth}
    \centering
    \caption*{\textsc{Crowd}}
     \includegraphics[width=1.0\textwidth]{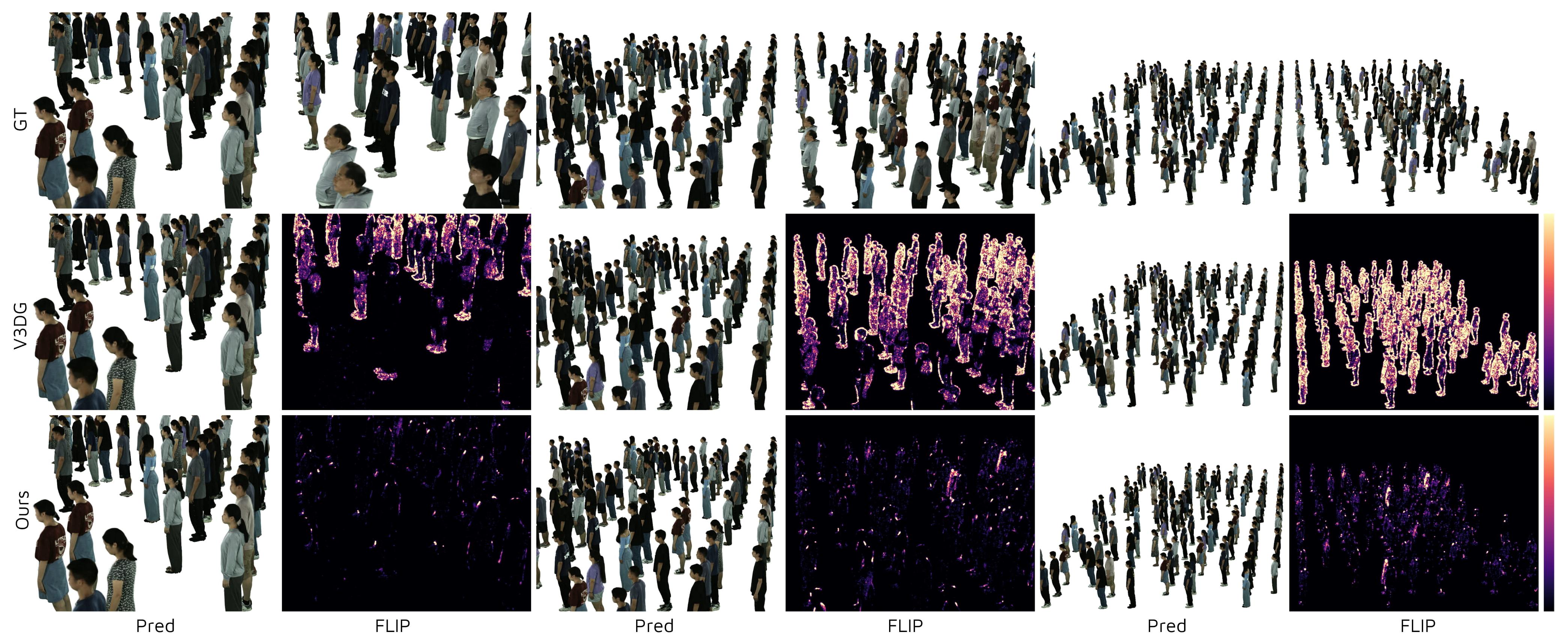}
  \end{subfigure}
  
  \caption{Shows GT render using gsplat, V3DG, and our method for close, medium, and far distances on the \textsc{DonutSea} and \textsc{Crowd} layouts. For V3DG and ours, we show the prediction on the left and FLIP on the right. We scaled FLIP by a factor of five to facilitate better visual comparison. These scenes have been downsampled by a factor of two compared to the original layouts provided by V3DG~\cite{Yang2025V3DG} to ensure that all methods can be tested without exceeding VRAM limits. }
  \label{fig:suppl_visual_comparison}
\end{figure*}

\begin{figure*}[htbp]
  \centering
   \begin{subfigure}[b]{1.0\textwidth}
    \centering
    \caption*{\textsc{Forest} (Large)}
     \includegraphics[width=1.0\textwidth]{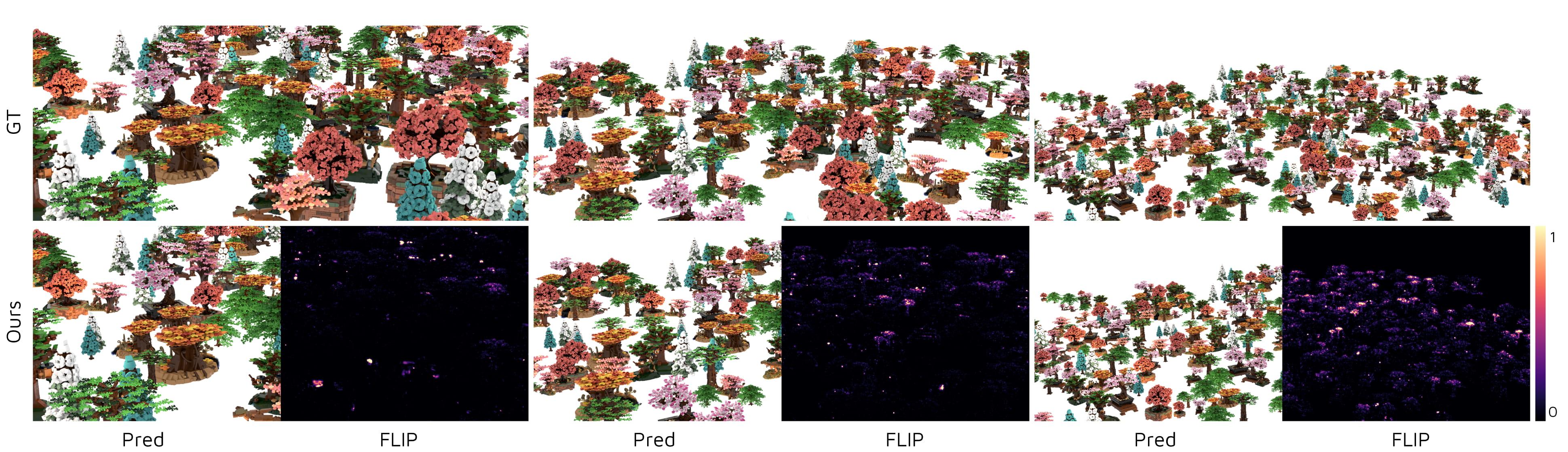}
  \end{subfigure}
  \begin{subfigure}[b]{1.0\textwidth}
    \centering
    \caption*{\textsc{DonutSea} (Large)}
     \includegraphics[width=1.0\textwidth]{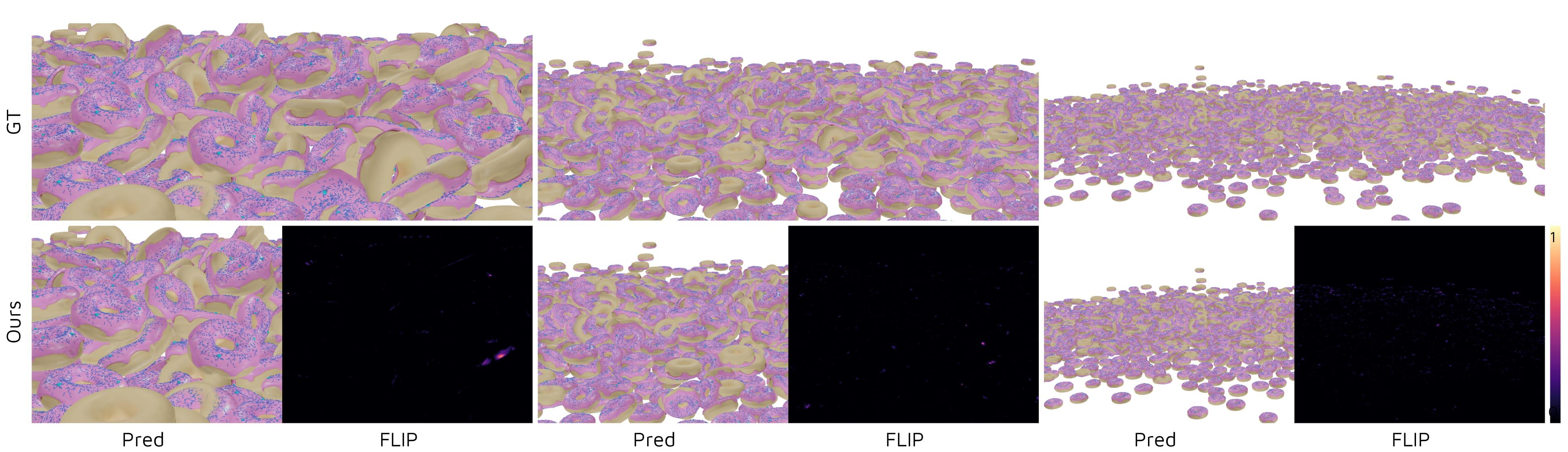}
  \end{subfigure}
  \begin{subfigure}[b]{1.0\textwidth}
    \centering
    \caption*{\textsc{Crowd} (Large)}
     \includegraphics[width=1.0\textwidth]{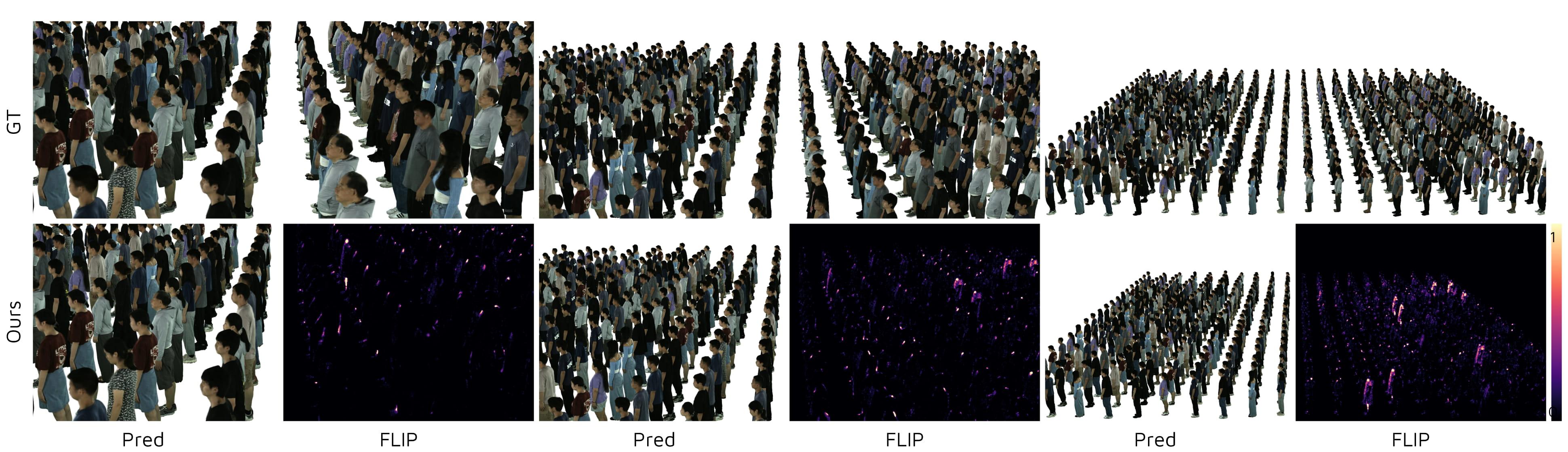}
  \end{subfigure}
  \caption{Shows GT render using our method with and without MLP(solely using instanced rasterizer) for close, medium, and far distances on the \textsc{Forest}, \textsc{DonutSea}, and \textsc{Crowd} layouts. For ours, we show the prediction on the left and FLIP on the right. These renderings show the original layout files without downscaling (individual assets remain centered after removing all Gaussians with an opacity lower than $\frac{1}{255}$).  We scaled FLIP by a factor of five to facilitate visual comparison.}
  \label{fig:large_scenes}
\end{figure*}

Our approach enables the rendering of large, composed scenes with significantly less VRAM than other methods, due to the combination of our instanced rasterizer and occlusion-culling MLP. In \cref{fig:suppl_visual_comparison}, we show the renders and FLIP metrics on the downscaled scenes compared to V3DG~\cite{Yang2025V3DG} for the scenes not in the main paper, and in \cref{fig:large_scenes}, we show the renders of the scene at full scale compared to our baseline. For both our method and V3DG, we observe a concentration of errors on the borders in the \textsc{Crowd} scene. While we the exact training setup of Yang~\etal~\cite{Yang2025V3DG} is not publicly available, we speculate that this may be due to their training on a background other than white. If this is the case, it can be solved by training on a random background, improving the results of both methods.
Our approach can run these large scenes on an NVIDIA RTX 3090 Ti with only 24 GB of VRAM at \textgreater20 FPS.

\subsection{Characterization of Impact Factors}
The impact of our approach on a given scene is influenced by several factors, such as the SH degree shown in \cref{fig:sh_comp}. 
We include this discussion to make these factors explicit and thus clarifying the strengths and limitations of our approach. 
%
% This analysis helps identify the specific scenarios in which our method has the greatest impact.

\paragraph{Compute per Gaussian}
Our approach scales effectively as the computational complexity per Gaussian increases.
This is a direct consequence of skipping the preprocessing of pruned Gaussians, which becomes more beneficial as per-Gaussian costs rise.
In \cref{fig:sh_comp}, we demonstrate this by increasing the degree of Spherical Harmonics (SH).
For future work involving more complex per-Gaussian computations, \eg, expensive shading/lighting routines, our approach provides a direct path for maintaining performance.

\paragraph{Gaussian/Asset Density}
As more Gaussians project to the same pixel, the standard tiled 3DGS rasterization algorithm becomes unable to balance loads effectively across tiles, which leads to reduced frame times.
NVGS helps in mitigating this by reducing the number of Gaussians that require processing.
While standard 3DGS was originally designed for spread distributions of Gaussians, the relative advantage of our approach grows in scenes with high Gaussian density, where assets are closely packed, and depth complexity per asset is high.

\paragraph{Screen Coverage}

Pruning Gaussians directly reduces the total number of generated Gaussian/tile instances.
Large Gaussians that overlap several tiles are the most impactful candidates for pruning, as a single visibility query can prevent redundant instancing across multiple tiles.
Conversely, for small Gaussians contained within a single tile, the relative computational savings from using NVGS are lower.

\paragraph{Occlusion Structure}

Since we do not frequency-encode the MLP inputs (which would help it to learn high-frequency visibility functions~\cite{tancik2020fourfeat}) to reduce computational demands, the occlusion structure of a scene influences performance.
To address this, we simplify the visibility function by smoothing it through auxiliary views and antialiasing.
While these simplifications aid convergence, they increase the number of Gaussians contributing to the final render, thereby reducing the number that can be pruned.
Consequently, scene parts with high-frequency visibility functions, such as a tree's crown, remain challenging for our method, as these simplifications increase the number of contributing Gaussians.

\paragraph{Distance to the Object}

The distance to the object also impacts performance. As our approach attempts to learn the asset's visibility as is, we are limited in pruning by how many contribute at far distances. Dedicated LoD approaches, however, can use fewer Gaussians to represent far distances. Compared to such methods (\eg., V3DG~\cite{Yang2025V3DG}), our methods perform well at close and medium distances, while they outperform us for far distances. 
For long distances, where these methods can use fewer Gaussians by merging them, we do not merge Gaussians.
The biggest jump is thus achieved up-close as can be seen in \cref{fig:graph_comparison}.

\begin{figure}[b]
  \centering
    \includegraphics[width=1\linewidth]{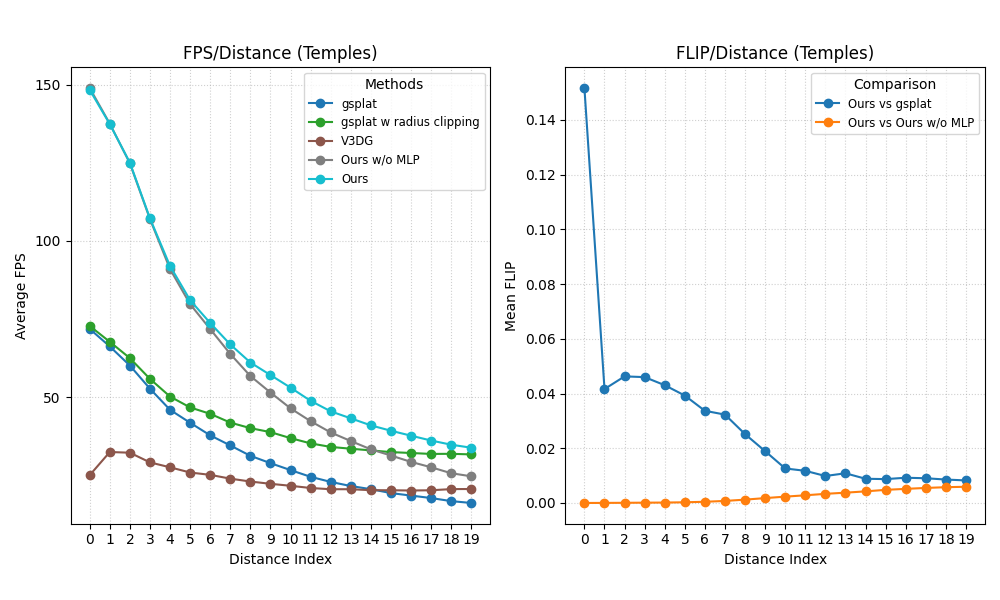}
  \caption{Average FPS/Distance plot for the \textsc{Temples} layout file. We additionally show the FLIP comparison against both the reference (gsplat) and ours w/o MLP. Our approach consistently achieves higher FPS, but due to floaters (\cref{fig:error_vis}), the metrics compared to the gsplat reference are surprisingly low. The FLIP comparison against ours w/o MLP shows that this is not a failure of our method but a difference in the rasterizer.}
  \label{fig:temple_fps}
\end{figure}

\subsection{More Detailed Objects}
% FPS/Distance
% FLIP/Distance 
% Explanation
% Large floater Gaussians that our formulation culls + image
% Doesn't meaningfully contribute to the image visually, small hue change
% shouldn't be there in the first place.

\begin{figure*}[b]
\centering
\includegraphics[width=0.68\linewidth]{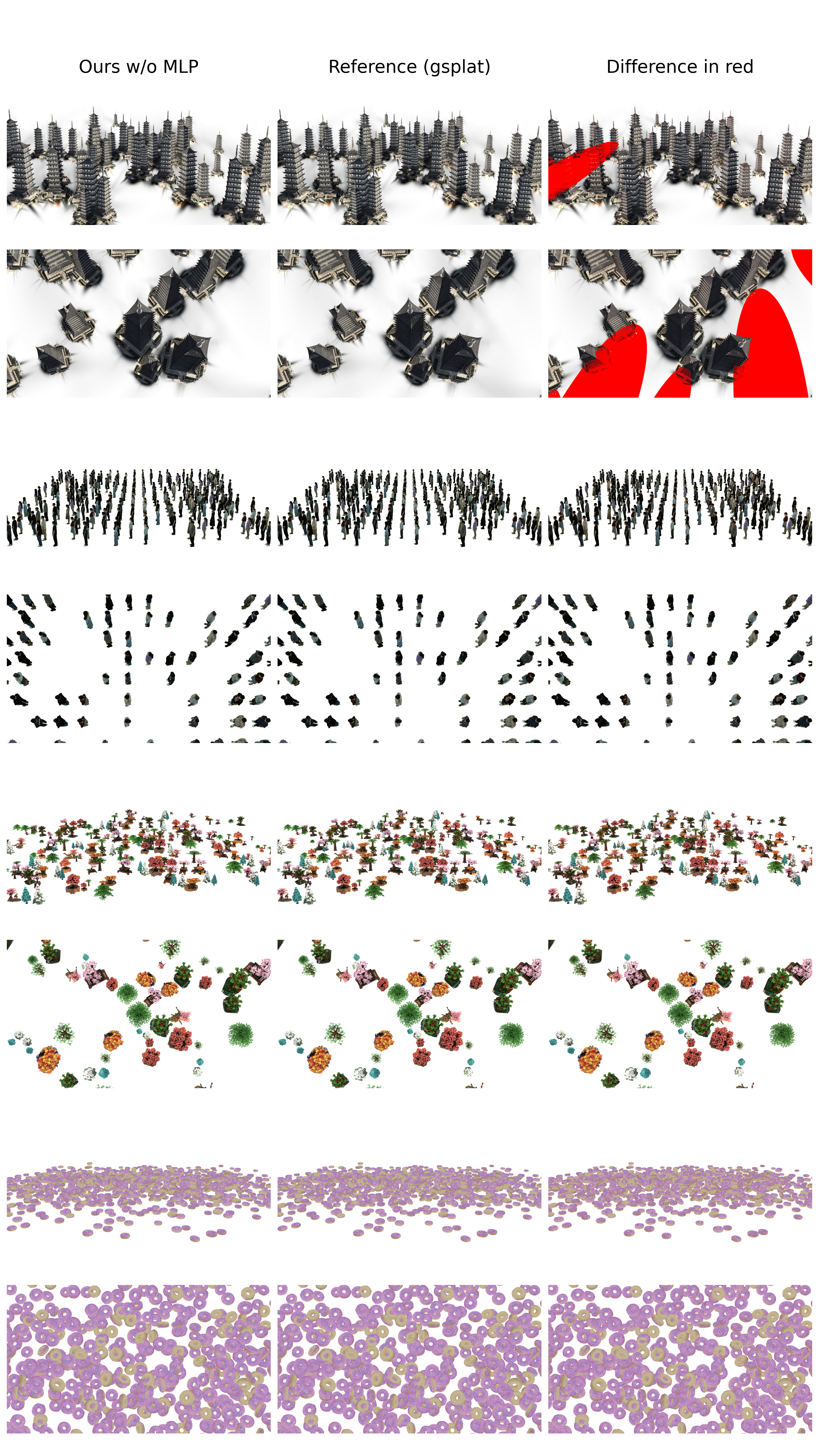}
\caption{Left: ours w/o MLP; center: reference rendering from gsplat; right: difference image where red pixels indicate any difference between left and center. For the \textsc{Temples} layout, there are large Gaussians that cover large parts of the scene, colored in red.
%We emphasize that the MLP is not used in these renders, so any differences result solely from changes to the underlying rasterizer.
While visually there is little to no difference, removing these Gaussians results in a small hue shift that significantly affects image metrics, especially PSNR.}
\label{fig:error_vis}
\end{figure*}

To evaluate scalability of our approach \wrt to asset complexity, we utilize the temple asset provided by Yang~\etal~\cite{Yang2025V3DG} and a downscaled version of their \textsc{Temples} layout.
The asset consists of approximately 1.3M Gaussians, which reduces to roughly 900k after standard opacity pruning ($<\frac{1}{255}$).
As shown in \cref{fig:temple_fps} and \cref{tab:temple_metrics}, our approach consistently maintains a lower VRAM footprint and higher FPS at close-to-medium distances.
While image metrics for this scene are lower when compared to the reference rendering from gsplat, visual inspection reveals that this is primarily due our rasterizer being able to cull large ``floaters'' present in the original asset more effectively (see \cref{fig:error_vis}).
This results in a better viewing experience and visual quality but does reduce image metrics.
We highlight that, as clearly shown in \cref{fig:temple_fps}, this behavior is independent of the visibility MLP predictions and solely depends on how the underlying rasterizer handles large splats.

\begin{table}[htb]
\caption{We show image metrics and VRAM usage for the \textsc{Temples} layout from V3DG. Our approach uses less VRAM and higher FPS for all distances in the layout. For this particular layout, we notice that image metrics are low at first glance. We show in \cref{fig:error_vis} that, although our approach visually appears the same, our rasterizer culling large Gaussians artificially deflates image metrics.}
\label{tab:temple_metrics}
\centering
\small
\setlength\tabcolsep{7.5pt}
\begin{tabular}{l|cccc}
\toprule
\textbf{Method} & PSNR$^\uparrow$ & SSIM$^\uparrow$ & FLIP$^\downarrow$ & VRAM$^\downarrow$ \\
\midrule
gsplat  & -- & -- & -- & ~~7.3GB\\
gsplat$^\dagger$  & 32.87 & 0.969 & 0.034 & ~~4.7GB\\
V3DG  & 52.30 & 0.998 & 0.004 & 10.3GB\\
Ours w/o MLP  & 45.21 & 0.995 & 0.027 & ~~4.2GB\\
Ours & 40.95 & 0.995 & 0.029 & ~~3.1GB\\
\bottomrule
\end{tabular}
\end{table}
\begin{table*}[b]
\centering
\begin{tabular}{lccccccr}
\toprule
\scriptsize
 & \multicolumn{3}{c}{\textbf{GT}} & \multicolumn{3}{c}{\textbf{Ours}} &  \\
\cmidrule(lr){2-4} \cmidrule(lr){5-7}
\textbf{Asset} & Passed & Used & Used (\%) & Passed & Used & Used (\%) & \textbf{$\Delta$ Passed (\%)} \\
\midrule
100831 & 313,257 & 151,748 & 48.4\% & 199,207 & 148,328 & 74.5\% & -36.41\% \\
100833 & 273,513 & 134,121 & 49.0\% & 183,906 & 130,578 & 71.0\% & -32.76\% \\
100835 & 302,469 & 153,949 & 50.9\% & 206,470 & 151,189 & 73.2\% & -31.74\% \\
100837 & 331,681 & 164,372 & 49.6\% & 214,328 & 159,772 & 74.5\% & -35.38\% \\
100839 & 295,192 & 146,948 & 49.8\% & 202,081 & 143,489 & 71.0\% & -31.54\% \\
100841 & 301,718 & 154,509 & 51.2\% & 194,411 & 151,447 & 77.9\% & -35.57\% \\
100843 & 278,513 & 141,775 & 50.9\% & 198,186 & 138,574 & 69.9\% & -28.84\% \\
100845 & 824,204 & 393,820 & 47.8\% & 494,592 & 387,509 & 78.3\% & -39.99\% \\
100847 & 334,039 & 158,388 & 47.4\% & 207,953 & 154,531 & 74.3\% & -37.75\% \\
100849 & 382,225 & 182,777 & 47.8\% & 235,006 & 177,868 & 75.7\% & -38.52\% \\
100851 & 318,754 & 149,637 & 46.9\% & 201,116 & 145,765 & 72.5\% & -36.91\% \\
100853 & 368,617 & 180,419 & 48.9\% & 241,633 & 176,145 & 72.9\% & -34.45\% \\
100855 & 286,111 & 140,272 & 49.0\% & 189,134 & 137,433 & 72.7\% & -33.90\% \\
100857 & 281,235 & 136,311 & 48.5\% & 181,743 & 132,082 & 72.7\% & -35.38\% \\
100859 & 229,308 & 109,789 & 47.9\% & 148,576 & 107,207 & 72.2\% & -35.21\% \\
100861 & 264,017 & 128,745 & 48.8\% & 170,439 & 125,885 & 73.9\% & -35.44\% \\
\midrule
bonsai & 319,339 & 209,366 & 65.6\% & 284,781 & 204,559 & 71.8\% & -10.82\% \\
crispy & 264,978 & 191,948 & 72.4\% & 237,725 & 191,353 & 80.5\% & -10.29\% \\
desk & 425,546 & 244,866 & 57.5\% & 339,417 & 237,974 & 70.1\% & -20.24\% \\
dinosaur & 221,706 & 144,505 & 65.2\% & 207,475 & 143,451 & 69.1\% & -6.42\% \\
fall & 276,574 & 147,584 & 53.4\% & 202,262 & 144,641 & 71.5\% & -26.87\% \\
house & 273,216 & 168,544 & 61.7\% & 240,408 & 167,239 & 69.6\% & -12.01\% \\
oak & 257,670 & 205,816 & 79.9\% & 246,392 & 205,118 & 83.2\% & -4.38\% \\
pine & 264,686 & 156,355 & 59.1\% & 202,004 & 152,392 & 75.4\% & -23.68\% \\
\midrule
donut & 44,106 & 26,267 & 59.6\% & 29,964 & 26,052 & 86.9\% & -32.06\% \\
\bottomrule
\end{tabular}
\caption{Per-asset statistics. We display the number of Gaussians passed to the rasterizer, as well as the Gaussians that actually contribute to the final rendering, both in absolute numbers and as percentages relative to the total number of Gaussians passed. We then show the difference in Gaussians passed between the ground truth and our method in the last column. We divide the assets into their three respective datasets. The first 16 entries are from the MVHumanNet dataset~\cite{xiong2024mvhumannet}, the next eight from the RTMV~\cite {tremblay2022rtmv} dataset, and the donut from~\cite{Vagadia2021Donut3D}. We prune a significant number of Gaussians across all assets. For oak and dinosaur, the lower gains can be dedicated to the higher-frequency visibility functions, which comprise the majority of the asset. While this could be improved by using frequency encodings, we argue that the slowdown in general for these specific assets is not worth the slight increase for a small subset of assets.}
\label{tab:gaussian_stats}
\end{table*}

% \begin{figure*}[b]
%     \centering
%     \includegraphics[width=1\linewidth]{figures/error_vis.png}
%     \caption{}
%     \label{fig:error_vis}
% \end{figure*}

\end{document}